# Migrating Birds Optimization-Based Feature Selection for Text Classification

Cem Kaya [1], Zeynep Hilal Kilimci [2], Mitat Uysal [3] and Murat Kaya [4]

[1] Scientific and Technical Research Council of Turkiye (TUBITAK) and Dogus University; cem.kaya@tubitak.gov.tr
[2] Department of Information Systems Engineering, Kocaeli University; zeynep.kilimci@kocaeli.edu.tr
[3] Department of Software Engineering, Dogus University; muysal@dogus.edu.tr
[4] Department of Computer Programming, Acibadem University; murat.kaya@acibadem.edu.tr
[*] Correspondence: zeynep.kilimci@kocaeli.edu.tr

**Abstract:** This research introduces a novel approach, MBO-NB, that leverages Migrating Birds Optimization (MBO) coupled with Naive Bayes as an internal classifier to address feature selection challenges in text classification having large number of features. Focusing on computational efficiency, we preprocess raw data using the Information Gain algorithm, strategically reducing the feature count from an average of 62221 to 2089. Our experiments demonstrate MBO-NB's superior effectiveness in feature reduction compared to other existing techniques, emphasizing an increased classification accuracy. The successful integration of Naive Bayes within MBO presents a well-rounded solution. In individual comparisons with Particle Swarm Optimization (PSO), MBO-NB consistently outperforms by an average of 6.9% across four setups. This research offers valuable insights into enhancing feature selection methods, providing a scalable and effective solution for text classification.

## 1. Introduction

Text classification is a type of supervised machine learning that automatically assigns predefined categories or labels to text documents [1–5]. The process involves several steps, including data preprocessing [6], feature selection [7], model training, and evaluation [8]. During data preprocessing, the input text is cleaned and transformed into a numerical representation, such as bag-of-words, TF-IDF, or word embeddings, to enable the use of machine learning algorithms. feature selection aims to capture the most important information from the text and remove irrelevant information. Model training involves selecting an appropriate algorithm, optimizing its parameters, and evaluating its generalization performance using a held-out set of labeled examples. Common machine learning algorithms used for text classification include Naive Bayes, Decision Trees, Support Vector Machines(SVM), and Neural Networks [9]. Text classification is an important problem in natural language processing with applications in various fields [10], such as sentiment analysis [11], spam detection [12], topic modeling [13,14], and language identification [1].

Feature selection is an important step in text classification where the goal is to identify the most relevant features that can be used to classify text documents into different categories [1,15]. In general, feature selection involves identifying a subset of features from a larger set of features that are considered to be most relevant for classification [10,16,17]. The goal of feature selection is to improve the accuracy of the classification model while reducing the computational complexity and the risk of over-fitting. There are several techniques for feature selection in text classification, including filter methods [18,19], wrapper methods [20,21], and embedded methods [22,23].

Heuristic optimization is another approach to solving optimization problems that is based on intelligent search and problem-solving techniques, rather than on mathematical models or exact algorithms [24–27]. Heuristic optimization algorithms are designed to explore the search space of a problem and find good solutions quickly, without necessarily guaranteeing that the solution found is optimal [28–31]. Heuristic optimization approaches typically involve generating a set of candidate solutions, i.e., neighbor states, evaluating their quality based on a fitness function or objective function, and then iteratively refining the solutions using some search or optimization method [32–34]. The search space may be explored using various techniques such as local search, simulated annealing [35], genetic algorithms, particle swarm optimization (PSO) [24,36], ant colony optimization [27], and many others [37–40]. Heuristic optimization approaches are often used in complex problems where exact solutions may be difficult to find, or where the search space is very large. Some examples of problems that can be tackled using heuristic optimization include vehicle routing [41], resource allocation, scheduling [42], network design [43], and many more [25,44–47].

The landscape of feature selection, particularly in scenarios with a large number of them, presented a challenge for conventional algorithms while highlighting the potential efficacy of heuristic approaches. The rationale behind this choice lies in the recognition of heuristics' prowess, particularly in situations where the search for global maxima becomes difficult, and local maxima suffice to address the complexities at hand. In this context, we leaned towards Migration Birds Optimization (MBO) [48], recognizing it as an emerging star in the field, offering efficient solutions to complex problems [49–52]. MBO is a nature-inspired optimization algorithm that is inspired from the flock of migrating birds. It mimics the collaboration and communication observed in bird flocks during migration, where individuals exchange information to enhance the overall group performance. In MBO, solutions are represented as individuals in a population, and the optimization process involves the iterative movement and interaction of these individuals to find optimal or near-optimal solutions to a given problem.

In this research, we propose a novel heuristic optimization based feature selection technique for text classification employing MBO [48]. To the best of our knowledge, this is the very first study that utilizes MBO as a feature selection algorithm in text classification. The contribution of this paper can be listed as follows:

- We propose a generic modeling approach, both for neighbor state and fitness function, which enables the utilization of various heuristic optimization techniques for feature selection.
- We use Information Gain as a baseline of our proposed approach enabling us to return the Information Gain data in the worst case scenario.
- The extensive experiments are carried out on widely-used data including 20News-18828, aahaber, hurriyet, mininews, and webkb4 in text classification domain.

The experiment results show that the inclusion of MBO technique as a feature selection model performs remarkable classification results when compared to state-of-the-art studies.

The rest of the paper is organized as follows: Section 2 presents a review of related work on different feature selection techniques. Section 3 details the proposed framework and methods used in this research. Section 4 gives the experimental setup and shares the results. Section 5 discusses the experimental results. Finally, Section 6 concludes the paper.

## 2. Related Work

MBO is a heuristic optimization algorithm that is inspired from the flock of migrating birds. It mimics the collaboration and communication observed in bird flocks during migration, where individuals exchange information to enhance the overall group performance. In MBO, solutions are represented as individuals in a population, and the optimization process involves the iterative movement and interaction of these individuals to find optimal or near-optimal solutions to a given problem. Due to its recent popularity, MBO has been applied to various real-world optimization problems such as land distribution problem [52], discrete problem [51], workers assignment balancing problem [53], flowshop scheduling problem [49,54,55], system identification problem [56], credit card fraud detection problem [57] and steel making-continuous casting problem [50].

In this section, a brief survey of state-of-the-art studies focused on the feature selection techniques are introduced. In a recent study ([19]), the authors introduce a novel multi-objective algorithm called Multi-Objective Relative Discriminative Criterion (MORDC) for feature selection in textual data. MORDC attempts to achieve a balance between minimizing redundant features and maximizing the relevance to the target class. In another recent study ([58]), the researchers introduce a novel multi-label feature selection method named MLACO based on Ant Colony Optimization. MLACO aims to identify the most promising features in the feature space by addressing both relevance and redundancy aspects. In [18], a suite of filter methodologies is systematically administered across synthetic data characterized by variations in the quantity of salient features, the magnitude of noise present in the output, the dynamic interplay among features, and a progressive augmentation in sample size. Information Gain [59] is another well-known feature selection method. Information Gain revolves around quantifying the relevance of features in a data by measuring their contribution to the reduction of uncertainty during the classification process. It finds application in a variety of research studies [59–61].

Particle Swarm Optimization (PSO) technique has been used widely in feature selection problem. One of the recent one is [62], that is a method that combines genetic algorithms and PSO are introduced to pick the best set of features. The main goal is to make it easier and faster to find the best solution for choosing features from large data. In another application of PSO ([36]), the authors present a novel approach to feature selection. It refines the problem by generating a subset of important features. This carefully selected subset of features demonstrates the effectiveness in improving how efficiently text classification can be applied, while also reducing the execution time. In a hybrid PSO algorithm with a smart learning approach [63], the researchers use a strategy that learns on its own to generate good options for exploration. At the same time, they use a competitive learning-based prediction strategy to make sure the algorithm is efficient in using the information it already has. This is done to balance between exploring new possibilities and exploiting the ones already known.

Ant Colony Optimization (ACO) has been used widely in feature selection problem. In a recent application of ACO [64], the researchers introduces a novel text feature selection approach employing a wrapper methodology, coupled with ACO to steer the feature selection process. Additionally, it employs the KNN as a classifier for assessing and generating a candidate subset of optimal features. The outcome of the feature subset, obtained through the proposed ACO-KNN algorithm, was utilized as input to identify and extract sentiment words from sentences in customer reviews. In [65], ACO for feature selection employs a hybrid search methodology that combine the benefits of both wrapper and filter approaches. To enable this hybrid search, they formulated novel sets of rules and heuristic information measurement. Simultaneously, the ants are directed toward accurate paths when constructing graph subsets, incorporating a bounded scheme at every step in the algorithm. Although this research aligns with our study in terms of heuristic application on feature selection problem, unfortunately, we could not find the tool to run the experiments.

## 3. Proposed Framework

MBO [48] is a population-based metaheuristic algorithm that mimics the social interaction and movement patterns observed in bird populations. MBO technique recognized for its efficacy in solving complex combinatorial problems [66–68]. In this research, we harness the inherent strengths of MBO to introduce a novel and promising MBO as a feature selection algorithm. In other words, the proposed approach capitalizes on MBO's ability to iteratively refine solutions, optimizing feature subsets to maximize classification performance.

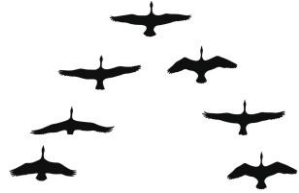

**Figure 1.** V-shaped formation of 7 birds.

This section provides a comprehensive overview of our novel approach, detailing the incorporation of MBO into the feature selection.

*3.1. Migration Birds Optimization*

The MBO model [48] is inspired by the behavior of bird flocks during migration, specifically $N$ birds, with $N$ being an odd number, flying in a *V*-shaped formation (Figure 1). Each bird represents a potential solution, undergoing refinement through the course of flights. A random bird is chosen as the leader, guiding the entire flock, while the remaining birds follow in two orderly lines, maintaining the *V*-shaped formation (Figure 1). This innovative approach leverages the dynamics of a bird flock to enhance the exploration and exploitation capabilities of the optimization algorithm.

Algorithm 1 outlines the MBO, which receives an input of $M \times N$ data and the number of birds in the flock, and strives to enhance the classification outcome by eliminating less significant features.

In the initial step, it calculates the fitness of the input data, which is a way to see how well the data works within the context of different solutions (is discussed in Section 3.2.3 in detail), denoted as $F$ where $0 \leq F \leq 1$ (line 1 of Algorithm 1). In line 2, we assign the initial fitness $F$ value to 3 different variables in order to observe the improvement of the fitness value. More specifically, $F_i$ ($1 \leq F \leq 3$) retains the fitness value from the fitness values of the iteration $i$ steps earlier (line 20). Hence, when the difference between $F_1$ and $F_3$ reaches 0 (line 8), it indicates that the same result has been achieved in the last three steps. At this point, we conclude the algorithm, presuming that no additional improvement can be achieved.

At line 6, the flock is initialized with an odd number of birds. An illustration of a flock with 7 birds is presented in Figure 1. The bird positioned at the forefront, i.e., at the corner of the *V*-formation, is designated as the leader of the flock. The flock iteratively engages in flight and landing cycles, wherein optimizations are applied to enhance the flock's performance. After completing a cycle, a new leader is chosen based on the bird with the highest fitness value, and the position of the current leader is replaced with the newly selected leader (line 19) while the positions of the remaining birds unchanged. Subsequently, the flock initiates another flight, following the same sequence of steps, and this process continues iteratively until the flock reaches its desired destination. This iterative process contributes to the ongoing refinement of the solutions explored by the bird flock.

We depict the steps carried out between lines 8-22 in Figure 2. Each step within the upper large box of Figure 2 corresponds to lines 11 in Algorithm 1. In these steps, we make some changes in the solutions which will be explained in detail in the following paragraph, with the goal of discovering more optimized solutions, referred to as birds, throughout the flight. The number of changes is decided by the number of *change* variable as given in line 9. For instance, if the *change* is 3, we make 3 changes in the current solution. If an optimization could have been made, the global best fitness value (lines 14-17) is updated. After 10 steps, a number determined through previous experiments, we reorder the flock by replacing the bird with the highest fitness value with the leader (line 19). The algorithm terminates when either no optimization is possible, i.e., the last three steps produces the same results, or it reaches max number of iterations, that is 100 (line 8). Then, the best solution found so far $F_{max}$ is returned as output.

**Algorithm 1** MBO algorithm
**input**: $MxN$ data, flockSize,
**output**: $M'xN$ data where $0 < M' \leq M$
1:  $F = computeFitness(data)$
2:  $F_1, F_2, F_3 \leftarrow F, F, F$
3:  $F_{max} \leftarrow F$
4:  $B_{max} \leftarrow data$
5:
6:  $flock \leftarrow initializeFlock(data, flockSize)$
7:  $counter \leftarrow 0$
8:  **while** $(counter < 3 \;||\; F1 - F3 == 0) \;\&\&\; (counter < 100)$ **do**
9:      $change \leftarrow calcChangeCount(count)$
10:     **for** $i$ in 1..10 **do**
11:         $flock.fly(M, change)$
12:         $Bird_{best} \leftarrow flock.findBestBird()$
13:         $F_{best} \leftarrow Bird_{best}.getFitness()$
14:         **if** $F_{best} > F_{max}$ **then**
15:             $F_{max} \leftarrow F_{best}$
16:             $B_{best} \leftarrow Bird_{best}$
17:         **end if**
18:     **end for**
19:     $flock.reorder()$
20:     $F_1, F_2, F_3 \leftarrow F_{max}, F_1, F_2$
21:     $counter \leftarrow counter + 1$
22: **end while**
23: **return** $B_{max}$

Each flock fly operation (line 10), i.e., shown as arrows in the Figure 2, corresponds to the operations given in Figure 3. We, for simplicity, elaborate a single iteration (line 10) with a hypothetical flock having 5 birds $b_1, b_2, b_3, b_4, b_5$ where $b_1$ is the leader, $b_2$ and $b_3$ follows the leader and $b_4$ and $b_5$ follows $b_2$ and $b_3$, respectively, in V-shaped formation as illustrated in Figure 3a. In Figure 3b, every bird $b_i$ generates a unique set of neighbors $B_i$ by creating similar solutions as itself through a small number of changes (Section 3.2.2). Then, in Figure 3c, the leader $b_1$ replaces itself with the best solution selected from the set $\{b_1\} \cup B_1$ and selects 2 more best solutions, i.e., the second $b_1''$ and the third best $b_1'''$
solutions, and $b_1''$ is added to its left set of birds and the $b_1'''$ is added to its right set of birds. Thus, the set of birds on the left side of the leader becomes $\{b_2\} \cup B_2 \cup b_1''$ and the set of birds on the right side of the leader becomes $\{b_3\} \cup B_3 \cup b_1'''$ (Figure 3c). In the fourth step (Figure 3d), the birds $b_2$ and $b_3$ replace themselves with the best solution selected from the sets $\{b_2 \cup B_2 \cup b_1''\}$ and $\{b_3 \cup B_3 \cup b_1'''\}$ respectively. Subsequently, both $b_2$ and $b_3$ selects their second best solutions, e.g., $b_2''$ and $b_3''$, and they are added to the left and right set of birds, respectively. In the last step, the birds $b_4$ and $b_5$ replaces themselves with the best solutions selected from the sets $\{b_4\} \cup B_4 \cup \{b_2''\}$ and $\{b_5\} \cup B_5 \cup \{b_3''\}$, respectively. Finally, the new flock contains $\{b_1', b_2', b_3', b_4', b_5'\}$ birds.

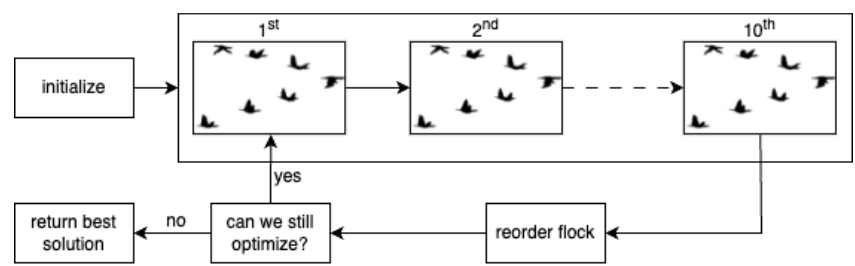

**Figure 2.** Overview of MBO algorithm.

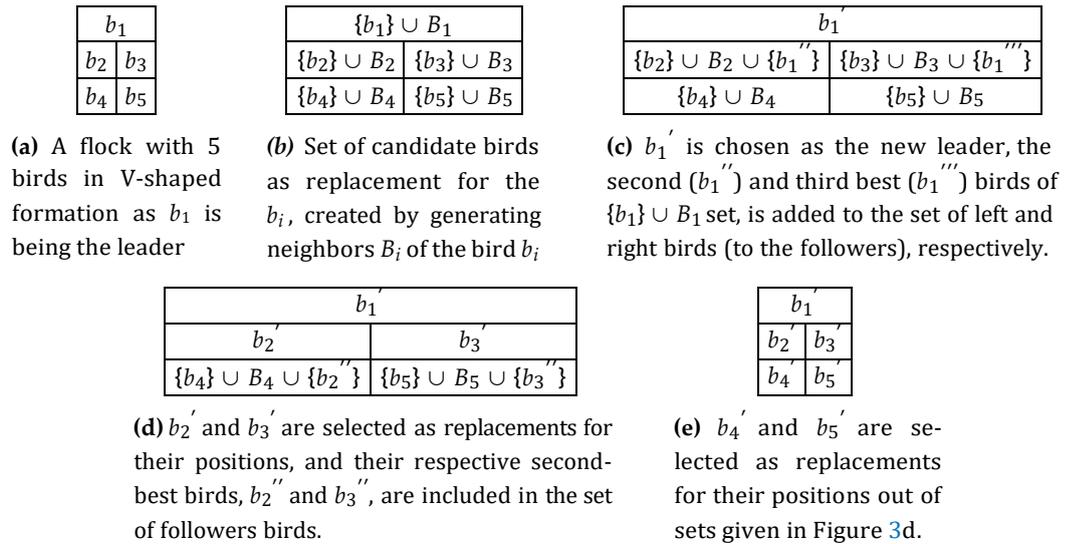

(a) A flock with 5 birds in V-shaped formation as $b_1$ is being the leader

(b) Set of candidate birds as replacement for the $b_i$, created by generating neighbors $B_i$ of the bird $b_i$

(c) $b_1{'}$ is chosen as the new leader, the second ($b_1{''}$) and third best ($b_1{'''}$) birds of $\{b_1\} \cup B_1$ set, is added to the set of left and right birds (to the followers), respectively.

(d) $b_2{'}$ and $b_3{'}$ are selected as replacements for their positions, and their respective second-best birds, $b_2{''}$ and $b_3{''}$, are included in the set of followers birds.

(e) $b_4{'}$ and $b_5{'}$ are selected as replacements for their positions out of sets given in Figure 3d.

**Figure 3.** An example of birds replacements in the flock by using neighbor states.

### 3.2. Modelling the Feature Selection as Heuristic Problem

In this section, we are not solely approaching the feature selection problem as an MBO problem. Instead, we present a generic modeling approach. This approach enables the utilization of various heuristic optimization techniques, provided they meet the specified requirements, for the feature selection.

Traditional heuristic methods typically require two key components: a *fitness function* and a *neighbor generation function*. The fitness function serves as a crucial evaluation metric, quantifying the quality or effectiveness of a potential solution within the problem space. It assigns a numerical value to each solution, indicating its performance with respect to the optimization goal. On the other hand, the neighbor generation function determines how neighboring solutions are generated. This function plays an important role in the exploration of the solution space, influencing the search strategy by defining the set of neighboring solutions considered for potential improvements. Together, the fitness function and the neighbor generation function form the core elements guiding the heuristic search process, allowing the algorithm to iteratively evaluate and refine solutions in the pursuit of optimal outcomes.

Therefore, we first introduce how we define a solution for feature selection problem and then present novel fitness and neighbor approaches for the problem in the following sections.

#### 3.2.1. Solution

The data is given in $NxM$ matrix format, where $N$ is the number of rows, e.g., textual data; news or tweets in this study, and $M$ is the number of features extracted from those texts. Based on this format, a *solution* $NxM'$ is represented as a subset of $M'$ features where $0 < M' < M$.

We represent all the solutions with an M-length vector with 0s and 1s such that $i$-th position of the vector represents whether the respective feature is chosen or not. For example, consider the vector given in Table 1 for hypothetical data having 10 features. This vector represents the reduced data that is constructed by selecting the features $f_0$, $f_1$, $f_4$, $f_7$, and $f_8$.

**Table 1.** A solution for a hypothetical data having 10 features.

| $f_0$ | $f_1$ | $f_2$ | $f_3$ | $f_4$ | $f_5$ | $f_6$ | $f_7$ | $f_8$ | $f_9$ |
|---|---|---|---|---|---|---|---|---|---|
| 1 | 1 | 0 | 0 | 1 | 0 | 0 | 1 | 1 | 0 |

**Table 2.** The corresponding reduced data for the solution given in Table 1.

|       | $f_0$    | $f_1$    | $f_4$    | $f_7$    | $f_8$    |
|-------|----------|----------|----------|----------|----------|
| $r_0$ | $v_{00}$ | $v_{01}$ | $v_{04}$ | $v_{07}$ | $v_{08}$ |
| $r_1$ | $v_{10}$ | $v_{11}$ | $v_{14}$ | $v_{17}$ | $v_{18}$ |
| $r_2$ | $v_{20}$ | $v_{21}$ | $v_{24}$ | $v_{27}$ | $v_{28}$ |
| .... |          |          |          |          |          |
| $r_N$ | $v_{N0}$ | $v_{N1}$ | $v_{N4}$ | $v_{N7}$ | $v_{N8}$ |

This reduced data, in fact, represents a $NxM'$ tabular data as given in Table 2.

This vector representation is chosen for storage efficiency not to copy the whole data repeatedly considering that thousands of solutions will be generated during the process.

3.2.2. Neighbor Generation

A neighbor state is generated by utilizing a solution and modifying it through the addition or removal of a small number of features. This operation is done randomly selecting positions in the feature vector and then applying the function given in (1) on each position. That is, converting 1s into 0s and 0s into 1. Therefore, a neighbor state can be considered as similar version of the state which it originates. This dynamic approach allows for the exploration of nearby solutions within the problem space. For instance, the feature vector given in Table 3, can be generated using the solution given in Table 1 by excluding the feature $f_0$ and including the feature $f_9$.

$$F(f) = 1 - f \tag{1}$$

**Table 3.** A neighbor state generated by excluding $f_0$ and including $f_9$ for the solution given in Table 1.

| $f_0$ | $f_1$ | $f_2$ | $f_3$ | $f_4$ | $f_5$ | $f_6$ | $f_7$ | $f_8$ | $f_9$ |
|-------|-------|-------|-------|-------|-------|-------|-------|-------|-------|
| **0** | 1     | 0     | 0     | 1     | 0     | 0     | 1     | 1     | **1** |

3.2.3. Fitness Function

We define the fitness function to assess the efficacy of the reduced data, quantifying it as the ratio of correctly classified instances. To do so, we employ the cross-validation technique using well-known Python library scikit-learn [69] to construct a classification model (Section 4.2) using the reduced data, i.e., the solution. Then, a superior solution is indicated by a higher ratio of correctly classified instances.

*3.3. MBO as Feature Selection*

MBO can be regarded as a conventional heuristic optimization technique, incorporating standard practices for fitness functions and neighbor utilization. Therefore, MBO is well-suited to serve as an effective approach for addressing the feature selection problem. This section outlines our utilization of MBO to tackle the feature selection problem.

We start with employing Info Gain [59,70] feature selection algorithm to reduce the feature set to a manageable size. This initial reduction is essential as MBO's efficiency is exponentially influenced by the number of features. The reduced data resulting from Info Gain serves as an input for MBO. Subsequently, we feed this reduced data into MBO, allowing heuristic operations to identify and select the most beneficial features among the remaining set. This approach ensures that, in the worst-case scenario, our selection aligns with the features identified by Info Gain. The Info Gain algorithm selects all features contributing to classification, unless the total selected features surpasses 2500 which is a limitation imposed for performance reasons of MBO. We, then, use the reduced data as input for MBO.

**Table 4.** Characteristics of the data used in the experiments.

| data | no of features | no of instances | no of classes | avg. word count per instance | avg. word length | language |
|---|---|---|---|---|---|---|
| 20News-18828 | 113241 | 18828 | 20 | 272.5 | 5.3 | English |
| aahaber | 48983 | 20000 | 8 | 33.6 | 6.5 | Turkish |
| hurriyet | 71071 | 6006 | 6 | 158.3 | 7.1 | Turkish |
| mininews | 35991 | 2000 | 20 | 311.0 | 6.3 | English |
| webkb4 | 41818 | 4199 | 4 | 281.0 | 6.3 | English |

We already explained the details of Algorithm 1 in Section 3. It takes this reduced $MXN$ data as input and produces a $M'XN$ solution by selecting $M'$ best features out of initial $M$ features.

**4. Experiments**

In this section, we evaluate the effectiveness of MBO in addressing the feature selection problem. The experiments are designed to compare its performance against well-known feature selection algorithms. For this purpose, in the first phase of experimentation, we attempt to select the best classifier to complement the MBO framework. This classifier selection is important because the fitness function (Section 3.2.3) determines the value of the solution. The computed fitness value, derived from this function, is subsequently employed to compare solutions and decide the best solution among others. In the second phase of experiments, we compare the performance of MBO with other well-known feature selection algorithms. The goal of these experiments is to figure out how well MBO can be applied in real-world problems by understanding its strengths and where it fits in the world of feature selection.

*4.1. Data*

In the experiments, we utilized a diverse set of data, each characterized by specific features, instances, and classes. The details of the data are outlined in Table 4. The table presents key information regarding five different data: 20News-18828 [71], aahaber [72], hurriyet [73], mininews [73], and webkb4 [74].

In Table 4, the "no of feature" column indicates the number of features or attributes associated with each data. The "no of instances" column denotes the total number of instances or data points in each data. The "no of classes" column represents the number of classes or categories present in each data. The "avg. word count per instance" column provides the average word count per data instance whereas the "avg. word length" column indicates the average length of words in each data. And lastly, the language column gives the language of the text used in the study. For instance, the data "hurriyet" contains 71071 features, 6006 instances, and 6 classes (or categories). The average word count per instance and average word length in data are 158.3 and 7.1, respectively. The text language is in Turkish.

We initially acquired the data in its raw textual format. During the data preprocessing phase, we employ tokenization to break down the text into individual units, followed by the removal of stop words to enhance the efficiency of subsequent analyses. Then, we utilized Term Frequency-Inverse Document Frequency (TF-IDF) [75] to evaluate the importance of terms in a document relative to a collection of documents. Finally, we construct feature vectors, representing the processed data in the feature vector format to further computational operations.

These data with varying characteristics provide a rich set of examples for our experiments, allowing us to explore and analyze the performance of different models across diverse data scenarios.

*4.2. Models*

In this section, the models employed in this research are briefly introduced. In the scope of this work, Information Gain [59], PSO [36] and MBO algorithms are assessed as feature selection techniques while Naive Bayes, Decision Tree, Multilayer Perceptron, and Support Vector Machine, are evaluated as machine learning techniques. In each model creation, use used 5-fold cross-validation technique.

The Naive Bayes algorithm, widely employed for classification tasks, operates based on the principles of Bayes' theorem [76,77], assuming conditional independence of features given the class label. The algorithm's functioning entails the preparation of labeled data for training, estimation of probabilities using Bayes' theorem, and prediction of class labels based on the highest probability.

Decision Tree algorithm [59,78] involves recursively partitioning the input data based on feature values to create a hierarchical structure resembling a tree. At each internal node of the tree, a decision rule is applied to determine the optimal feature to split the data, maximizing the homogeneity within each resulting subset. The process continues until a stopping criterion is met, typically when the data subsets reach a predetermined purity threshold or a specified depth is reached. The resulting tree can then be used to make predictions for new, unseen instances by traversing the tree from the root node to a leaf node, following the decision rules at each internal node.

In Multilayer Perceptron (MLP), information moves forward from one layer to the next, starting with the input nodes, then progressing through hidden layers (which can be one or more layers), and finally reaching the nodes in the output layer [79]. These networks are made up of a mix of simpler models known as sigmoid neurons. MLPs consist of many nodes, and sigmoidal functions are commonly used as activation functions. MLPs have the ability to grasp the complex and nonlinear decision boundaries from data. They typically have one input layer, one output layer, and multiple hidden layers positioned between the input and output layers. The inclusion of more hidden layers enables MLPs to understand more intricate nonlinear relationships between the input and output.

Particle Swarm Optimization (PSO), introduced by Kennedy and Eberhart in 1995 [24], represents an emergent population-based meta-heuristic. This method emulates social behaviors, specifically the phenomenon of birds flocking towards a promising position to achieve precise objectives within a multidimensional space. Similar to evolutionary algorithms, PSO conducts searches utilizing a population, referred to as a swarm, composed of individuals termed particles. These particles undergo iterative updates to enhance their positions. In the pursuit of discovering the optimal solution, each particle adjusts its search direction based on two influential factors: its individual best previous experience and the collective best, incorporating both cognitive and social aspects. It is characterized by its straightforward implementation requiring few parameters, and it has found extensive application in addressing optimization problems, including those related to feature selection [80].

Support Vector Machines algorithm [81,82] involves identifying an optimal hyperplane in a high-dimensional feature space to separate different classes of data points. SVM achieves this by maximizing the margin, which is the distance between the hyperplane and the nearest data points from each class, known as support vectors. This optimal hyperplane is determined by solving a convex optimization problem that aims to minimize classification errors while maximizing the margin.

Information Gain revolves around quantifying the relevance of features in a data by measuring their contribution to the reduction of uncertainty during the classification process [59,60]. It calculates the information gain by evaluating the difference in entropy (or impurity) before and after the feature is considered. Features with higher information gain are deemed more informative and are thus selected for further analysis.

*4.3. Results*

In this section, we share the experimental results. In the first phase, we aimed to iden- tify the most suitable classification algorithm for integration into our proposed approach. For this purpose, we employed a set of diverse classifiers, namely Decision Tree, Multilayer Perceptron, Naive Bayes, and Support Vector Machine. The objective of this phase was to identify the optimal classifier that aligns most effectively with the requirements of our algorithm. However, we did not choose any complex classification algorithms that require long running time such as Deep Learning techniques [7,73] since our proposed approach needs to create classification models repeatedly.

**Table 5.** Comparing classifier integrations of MBO in terms of correctly classified percentage of the reduced data.

| data | decision tree | multilayer perceptron | naive bayes | support vector machines |
|---|---|---|---|---|
| 20News-18828 | 4.4 | 47.4 | 83.0 | 5.3 |
| aahaber | 32.2 | 76.1 | 79.7 | 44.0 |
| hurriyet | 43.0 | 62.0 | 76.4 | 42.7 |
| mininews | 30.1 | 69.1 | 85.5 | 48.2 |
| webkb4 | 76.9 | 80.1 | 91.1 | 39.1 |

Note that the average feature count across the data (Table 4) is approximately 62221. Given the significant impact of feature count on the performance of our fitness function, we took measures to enhance efficiency. To do so, we applied the Information Gain algorithm to the raw data before starting the MBO algorithm. That is, the input data for the MBO algorithm is derived from the output of the Information Gain process. To mitigate computational demands, we imposed a cap on the maximum number of features, limiting it to 2500. It is important to emphasize that, for efficiency considerations, we utilized Information Gain (Section 4.2) as a representative of our raw data, rather than processing the entire data directly. This strategic approach aims to balance computational efficiency with the essential need to optimize the performance of our fitness function.

We share the first set of results in Table 5. For each data, we applied MBO with different classifiers, reporting the correctly classified ratio of final reduced data. We applied a 10-hours threshold to ensure timely completion of the experiments, while also accounting for practical considerations and constraints. Therefore, if an experiment exceeded this threshold, the best result computed within the timeframe is reported.

In Table 5, results from 20 different configurations, involving 4 classifiers and 5 data, are presented. The overall average correctly classified percentages are 37.3, 66.9, 83.1, and 35.9 for Decision Tree, Multilayer Perceptron, Naive Bayes, and Support Vector Machines, respectively. Notably, the highest percentages are observed for Multilayer Perceptron and Naive Bayes. To further investigate, we compared their results individually. Specifically, the minimum and the maximum correctly classified percentages is 47.4 and 80.1 for Multilayer Perceptron, respectively and 76.4 and 91.1 for Naive Bayes, respectively.

Based on the result of the initial set of experiments, we have chosen to incorporate Naive Bayes as the classifier for the fitness function within the MBO algorithm. Henceforth, we refer to this integrated approach as MBO-NB.

In the second set of experiments, we attempted to compare the effectiveness of MBO-NB's output data with its raw data, and reduced data by Information Gain and PSO. For a fair comparison, our envisioned approach is as follows: users initially experiment with several classifiers using a 5-fold cross-validation. Subsequently, only the best-performing classifier is selected and applied to the test data. Thus, our reporting focuses solely on the results obtained using the best classifier out of Naive Bayes, Multilayer Perceptron and Decision Tree. However note that this is for the evaluation of the reduced data, MBO-NB still uses Naive Bayes as internal classifier.

We present these results in Table 6 by marking the ones with the highest score among others with an asterisk "*". Note that PSO could not succeed to reduce the data within the time limitation. Therefore, to make a fair comparison, we exclude the result for aahaber and present the statsitics for the reamining data. That is, for instance, taking average of 20News-18828, hurriyet, mininews and webkb4 results. With this comparison, the average percentage for the comparable results are 84.8, 83.7, 87.6, and 80.4, for the raw data, Information Gain, MBO-NB and PSO, respectively. In order to use all data results in the comparison, we exclude the PSO results and take the averages of all data results. The averages become 84.1, 82.9, and 86.7, for the raw data, Information Gain, and MBO-NB respectively. In both comparisons, except 20News-18828 data, MBO-NB outperforms better than other techniques, that is 80% of the setup. More precisely, MBO-NB improved the percentages by 3.3%, 4.6%, and 9.2%, compared to the raw data, Information Gain and PSO, respectively.

**Table 6.** Comparing correctly classified percentages of the raw data and reduced data by Information Gain, MBO-NB and PSO.

| data | raw data | information gain | MBO-NB | PSO |
|---|---|---|---|---|
| 20News-18828 | *86.3 | 81.5 | 83.0 | 75.6 |
| aahaber | 81.0 | 79.7 | *82.8 | - |
| hurriyet | 74.5 | 76.4 | *82.0 | 72.5 |
| mininews | 87.8 | 85.5 | *94.2 | 86.0 |
| webkb4 | 90.7 | *91.3 | *91.3 | 87.5 |

**Table 7.** Comparing the feature counts of the raw data and reduced data by Information Gain, MBO-NB and PSO.

| data | raw data | information gain | MBO-NB | PSO |
|---|---|---|---|---|
| 20News-18828 | 113241 | 2500 | 1847 | 1268 |
| aahaber | 48983 | 2500 | 1696 | - |
| hurriyet | 71071 | 2500 | 1352 | 967 |
| mininews | 35991 | 467 | 193 | 224 |
| webkb4 | 41818 | 2422 | 1484 | 1153 |

Additionally, we share the feature counts for both the raw data and the reduced data resulting from the feature selection algorithms in Table 7. On average, Information Gain reduced the number of features in the raw data from 62221 to 2089, accompanying a decrease of 1.3 percentages in the correctly classified percentages. This trade-off is made to address performance concerns, meaning that we begin with a less favorable initial data, that is reduced data by Information Gain, but achieve a better correctly classified percentage compared to raw data , i.e., 3.3%.

## 5. Discussions

Given the substantial feature count, on average 62221, and its impact on our fitness function performance, we preprocessed the raw data using the Information Gain algorithm. This reduced the number of features from 62221 to 2089 on average, with a cap set at 2500 for efficiency. Despite using Information Gain as a representative of raw data for efficiency considerations, our strategic approach balances computational efficiency with the essential need to optimize our fitness function's performance.

Our experiments shows the effectiveness of MBO-NB, emphasizing its superiority in feature reduction over other techniques. The strategic integration of Naive Bayes as the internal classifier within MBO proves successful, offering a balanced solution for enhancing computational efficiency while maintaining optimal classification accuracy.

We also conduct an individual comparison of MBO and PSO, both utilizing heuristic approaches. Across all 4 setups, MBO-NB consistently outperforms PSO by an average of 6.9%.

## 6. Conclusions

We evaluated the effectiveness of the Migration Birds Optimization algorithm for the feature selection problem. Through a series of carefully designed experiments, we explored the performance of MBO in comparison to established feature selection algorithms, with a particular focus on classifier selection and subsequent assessments.

The initial phase of experimentation led us to the integration of Naive Bayes as the fitness function classifier within the MBO algorithm. This decision was informed by an analysis of the results obtained from different classifiers in terms of classification accuracy. The selection of Naive Bayes aligns with the iterative nature of our proposed approach, emphasizing the need for repeated creation of classification models.

Subsequent experiments, comparing MBO-NB with raw data, Information Gain, and Particle Swarm Optimization (PSO), revealed promising outcomes. MBO-NB consistently outperformed raw data and Information Gain, demonstrating its robustness in feature selection tasks. The comparative analyses showcased its superiority over PSO which is also a heuristic approach, reinforcing the efficacy of the proposed approach.

Furthermore, the analysis of the relationship between raw data and Information Gain-reduced data highlighted the potential of MBO-NB to enhance accuracy, presenting a viable alternative for real-world applications.

In summary, the results of this research emphasize the effectiveness of MBO-NB as a feature selection algorithm. The careful selection of classifiers, coupled with strategic preprocessing steps, contributes to the algorithm's efficiency and performance. The findings presented in this study provide valuable insights into the application of MBO-NB in diverse scenarios, with implications for data-driven decision-making processes.

As we move forward, future research endeavors should thoroughly explore specific domains, examining the adaptability of MBO-NB to varying data and problem contexts. The insights gained from this research contribute to the changing nature of feature selection methodologies, opening avenues for further exploration and refinement. Furthermore, another potential area for future investigation could be the integration of Deep Learning algorithms and text-based transformer approaches within the MBO framework.